\documentclass{article}

\usepackage{times}
\usepackage{graphicx} 
\usepackage{multirow}
\usepackage{amsmath}

\usepackage{amssymb}

\usepackage{enumitem} 

\usepackage{hyperref}

\usepackage[accepted]{icml2016}

\icmltitlerunning{GPU Activity Prediction using Representation Learning}

\pdfoutput=1

\begin{document} 

\twocolumn[
\icmltitle{GPU Activity Prediction using Representation Learning}

\icmlauthor{Aswin Raghavan, Mohamed Amer, Timothy Shields, David Zhang, Sek Chai}{firstname.lastname@sri.com}
\icmladdress{SRI International,
	201 Washington Rd. Princeton, NJ08540}
\icmlkeywords{Representation learning; Time-Series analysis; Cognitive Architecture}

\vskip 0.3in
]

\begin{abstract} 
GPU activity prediction is an important and complex problem. This is due to the high level of contention among thousands of parallel threads. This problem was mostly addressed using heuristics. We propose a representation learning approach to address this problem. We model any performance metric as a temporal function of the executed instructions with the intuition that the flow of instructions can be identified as distinct activities of the code. Our experiments show high accuracy and non-trivial predictive power of representation learning on a benchmark.
\end{abstract} 
\section{Introduction}\label{sec:Intro}
The performance of a computing system relies on the sustained operational throughput. Sustained operation is becoming harder to achieve as computation workloads become more complex. At the same time, with the end of Dennard scaling \cite{Esmaeilzadeh_ISCA2011}, and the increasing abundance of Big Data, it is imperative to minimize wasted processor effort in order to achieve processor reliability and scalability \cite{Wulf_ACMSIG1995, Patterson_BEARS2006, Kogge_TechReport2008}.

The goal of this paper is to demonstrate the efficacy of machine learning to designing computing systems. We anticipate that classical problems such as branch predictions and cache management can be re-evaluated such that heuristically based approaches \cite{Yeh_ISM1991, fung2007dynamic,  li2015priority} can be replaced with a machine learning approach \cite{Leng_ISM2015}. It is well understood in the computer architecture community that processor behavior is highly complex and data dependent. Processor data is widely available in the form of benchmarks \cite{che2010characterization}, and algorithms are extensively compared using these benchmarks \cite{blem2011challenge}.

We choose a well-understood and well-defined problem of predicting GPU Cache Misses \cite{li2015priority,chen2014adaptive} for this paper. General Purpose GPU (GPGPU) achieve high throughput execution via a high level of parallelism. Predicting GPU Cache Misses is complex due to the high level of contention among thousands of threads. Cache contention is a bottleneck for parallel execution when many threads are waiting for cache operation, causing the addition of more threads (or cores) to be detrimental. Predicting whether a cache miss is about to occur is useful for better cache management such as cache bypassing \cite{chen2014adaptive}, pre-fetching \cite{lee2010many}, prioritized allocation \cite{li2015priority} etc. Further, cache misses indirectly cause increased energy and power usage \cite{Leng_ISM2015} because of second order effects beyond memory latency. In principle, our approach is amenable to predict these higher order events (such as voltage scaling \cite{Leng_ISM2015} and faults \cite{Chai_WNTC2014}) either directly \cite{tiwari1994power} or via hierarchical modeling.

We propose a new model that can predict key processor events that limit processor throughput. 
We propose a new variant of the Conditional Restricted Boltzmann Machines (CRBMs) \cite{Taylor_JMLR2011} to directly address system performance and reliability. 
CRBMs efficiently model short-term temporal phenomenon. 
Prior work used a perceptron to predict cache misses \cite{leng2013gpuwattch}. Unlike their approach, our model accounts for time-series and count data. 

Our approach assumes the availability of a simulator for CUDA \cite{bakhoda2009analyzing} that can generate a dataset for training our model. In principle, this approach can be used in real-time by incrementally augmenting the dataset. Multiple repeated executions can even lead to increased predictive power because more data is available for machine learning. 
Our predictor is naturally agnostic to the hardware and architecture as it relies on execution traces. 

{\it Our contributions:} 
\begin{itemize}[itemsep=-1pt,topsep=-2pt, partopsep=1pt]
	\item Prediction of processor events as temporally-extended activities in a stream of instructions.
	\item Using Discriminative Conditional Restrictive Boltzmann Machines (DCRBM) to learn processor states.
\end{itemize}
\section{Literature Review}\label{sec:LitReview}
\noindent{\bf Cache Miss Prediction:} There is a large body of research on branch prediction to improve cache performance. Simple static solutions can achieve 80\% correct prediction by analyzing control-flow and static heuristics \cite{Ball_ACM1993}. Dynamic solutions \cite{Yeh_ISM1991} are more complicated as they are implemented with counters and tables to store branch history based on branch memory address. Other approaches that are data-driven use perceptrons \cite{Jimenez_HPCA2001} and feed-forward neural networks \cite{Calder_TOPLAS1997}. 

\noindent{\bf Representation Learning:} Restricted Boltzmann Machines (RBMs) form the building blocks in energy based deep networks \cite{Hinton_NC2006, Salakhutdinov_Science2006}. Recently, temporal models based on deep networks have been proposed, capable of modeling a more temporally rich set of problems. These include Conditional RBMs (CRBMs) \cite{Taylor_JMLR2011} and Temporal RBMs (TRBMs) \cite{Sutskever_AISTATS2007}. CRBMs have been used in both visual \cite{Taylor_JMLR2011} and audio \cite{Mohamed_ICML2009}. In addition to efficiently modeling time-series data, RBMs were formulated to be trained discriminatively for classification \cite{Larochelle_ICML2008}, and model word-count vectors from a large set of documents \cite{Salakhutdinov_IJAR2009}.
\section{Model}\label{sec:Model}
The input to our model (called visible units) is an instruction mix per time step, ie.\! the histogram of counts of instructions being excuted, obtained from the GPU simulator. The labels are any chosen performance metric also output by the simulator.

We discuss a sequence of models, gradually increasing in complexity, so that the different components of our model can be understood in isolation. We start with the basic CRBM model, then we extend to the discriminative DCRBM, and finally CountDCRBM. 

\noindent{\bf Conditional Restricted Boltzmann Machines:} CRBMs \cite{Taylor_JMLR2011}, are a natural extension of RBMs for modeling short term temporal dependencies. A CRBM is an RBM which takes into account history from the previous time instances $t-N,\hdots,t-1$ at time $t$. This is done by treating the previous time instances as additional inputs. Doing so does not complicate inference. ${\bf v}$ is a vector of visible nodes, ${\bf h}$ is a vector of hidden nodes, and ${\bf v}_{<t}$ is the visible vectors from the previous $N$ time instances, which influences the current visible and hidden vectors. $E_{\text{C}}$ is the energy function, and $Z$ is the partition function. The parameters ${\boldsymbol \theta}$ to be learned are ${\bf a}$ and ${\bf b}$ the biases for ${\bf v}$ and ${\bf h}$ respectively and the weights ${\it W}$. $A$ and $B$ are matrices of concatenated vectors of previous time instances of ${\bf a}$ and ${\bf b}$. The CRBM is fully connected between layers, with no lateral connections.  This architecture implies that {\bf v} and {\bf h} are factorial given one of the two vectors.  This allows for the exact computation of $p_{\text{C}}({\bf v}|{\bf h},{\bf v}_{<t})$ and $p_{\text{R}}({\bf h}|{\bf v},{\bf v}_{<t})$.  Some approximations have been made to facilitate efficient training and inference, more details are available in \cite{Taylor_JMLR2011}. A CRBM defines a probability distribution $p_{\text{C}}$ as a Gibbs distribution (\ref{eqn:CRBM}). 
\begin{figure}[t]
	\centering
	\includegraphics[width=0.65\columnwidth]{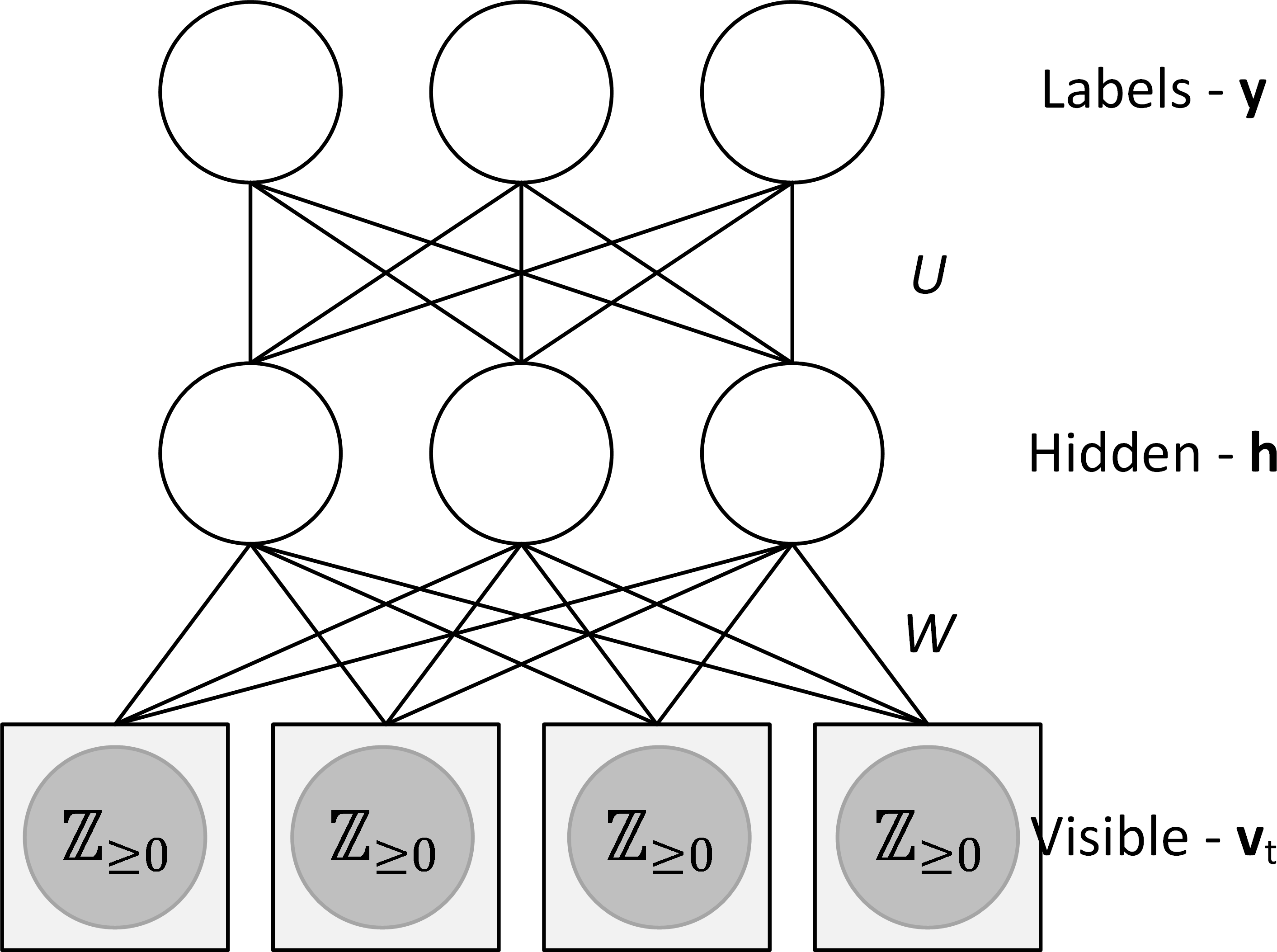}	
	\caption{This figure illustrates the CountDCRBM model.}
	\label{fig:ModelFigure}
\end{figure}
\begin{equation}
\centering
\begin{array}{c}
p_{\text{C}}({\bf v}_{t},{\bf h}_{t}|{\bf v}_{<t})=\exp[-E_{\text{C}}({\bf v}_{t},{\bf h}_{t}|{\bf v}_{<t})]/Z({\boldsymbol \theta}).
\end{array}
\label{eqn:CRBM}
\end{equation}
The energy function $E_{\text{C}}({\bf v}_{t},{\bf h}_{t}|{\bf v}_{<t})$ in (\ref{eqn:ECRBM}) is defined in a manner similar to that of the RBM.
\begin{equation}
\resizebox{0.9\linewidth}{!}{$
\begin{array}{ll}
E_{\text{C-Real}}({\bf v}_{t},{\bf h}_{t}|{\bf v}_{<t})=&-\sum_{i} (c_{i}-v_{i,t})^2/2 - \sum_{j} d_{j} h_{j,t}\\
&- \sum_{i,j} v_{i,t} w_{i,j} h_{j,t},\\
E_{\text{C-Binary}}({\bf v}_{t},{\bf h}_{t}|{\bf v}_{<t})=&-\sum_{i} c_{i}v_{i,t} - \sum_{j} d_{j} h_{j,t}\\
&-\sum_{i,j} v_{i,t} w_{i,j} h_{j,t}, \\
E_{\text{C-Count}}({\bf v}_{t},{\bf h}_{t}|{\bf v}_{<t})=&-\sum_{i} (c_{i}v_{i,t} - \log(v_{i,t}!)) \\
&- \sum_{j} d_{j} h_{j,t} - \sum_{i,j} v_{i,t} w_{i,j} h_{j,t},
\end{array}$}
\label{eqn:ECRBM}
\end{equation}
The probability distributions for the visible nodes are defined in (\ref{eqn:PCRBMV}), 
\begin{equation} 
\resizebox{0.9\linewidth}{!}{$
\begin{array}{l}
p_{\text{C-Real}}(v_{i,t}|{\bf h}_{t},{\bf v}_{<t})	= \mathcal{N}(c_{i}+ \sum_{j} h_{j,t}w_{i,j},1),\\
p_{\text{C-Binary}}(v_{i,t} = 1|{\bf h}_{t},{\bf v}_{<t}) =	\sigma(c_{i}+ \sum_{j} h_{j,t}w_{i,j}),\\
p_{\text{C-Count}}(v_{i,t}|{\bf h}_{t},{\bf v}_{<t}) = \mathcal{P}(m, \exp(c_i + \sum_{j} h_{j}w_{ij})), 
\end{array} $}
\label{eqn:PCRBMV}
\end{equation}
where, $\mathcal{N}$ is a normal distribution, $\sigma$ is a sigmoid distribution, and $\mathcal{P}$ is a Poisson distribution. The hidden nodes is defined in (\ref{eqn:PCRBMH}),
\begin{equation}
p_{\text{C}}(h_{j,t} = 1 |{\bf v}_{t},{\bf v}_{<t})	= \sigma(d_{j} + \sum_{i} v_{i,t} w_{i,j}).
\label{eqn:PCRBMH}
\end{equation}
%
%
\begin{equation}
c_{i} = a_{i} + \sum_{p}A_{p,i} v_{p,<t}, \quad d_{j} = b_{j} + \sum_{p}B_{p,j} v_{p,<t}.
\label{eqn:CDCRBM}
\end{equation}
\noindent{\bf Discriminative CRBMs:} DCRBMs are based on the model in \cite{Larochelle_ICML2008}, generalized to account for temporal phenomenon using CRBMs. DCRBMs are a simpler version of the Factored Conditional Restricted Boltzmann Machines \cite{Taylor_JMLR2011} and Gated Restricted Boltzmann Machines \cite{Memisevic_CVPR2007}. Both these models incorporate labels in learning representations, however, they use a more complicated potential which involves three way connections into factors. DCRBMs define the probability distribution $p_{\text{DC}}$ as a Gibbs distribution (\ref{eqn:DCRBM}). 
\begin{equation} \label{eqn:DCRBM}
\resizebox{0.9\linewidth}{!}{$
p_{\text{DC}}({\bf y}_{t},{\bf v}_{t},{\bf h}_{t}|{\bf v}_{<t};{\boldsymbol \theta})=\frac{1}{Z({\boldsymbol \theta})}\exp[-E_{\text{DC}}( {\bf y}_{t},{\bf v}_{t},{\bf h}_{t}|{\bf v}_{<t})]
$}
\end{equation}
The hidden layer ${\bf h}$ is defined as a function of the labels $y$ and the visible nodes ${\bf v}$. A new probability distribution for the classifier is defined to relate the label $y$ to the hidden nodes ${\bf h}$ as in (\ref{eqn:PDCRBMH}), 
as well as relate ${\bf h}$ to $y$ as in (\ref{eqn:PDCRBMY}). The new energy function $E_{\text{DC}}$ is shown in (\ref{eqn:EDCRBM}).
\begin{equation} \label{eqn:PDCRBMH}
\resizebox{0.9\linewidth}{!}{
$p_{\text{DC}}(h_{j,t} = 1 |y_{t},{\bf v}_{t},{\bf v}_{<t})= \sigma(d_{j}+ u_{j,k} + \sum_{i} v_{i,t} w_{ij})$,}
\end{equation}
\begin{equation} \label{eqn:PDCRBMY}
p_{\text{DC}}(y_{l,t}|{\bf h}_{t})=\frac{\exp[s_{l}+\sum_j u_{j,l}h_{j,t}]}{\sum_{l^*}\exp[s_{l^*}+\sum_j u_{j,l^*}h_{j,t}]}
\end{equation}
\begin{equation} \label{eqn:EDCRBM}
\resizebox{0.9\linewidth}{!}{
$E_{\text{DC}}({\bf y}_{t},{\bf v}_{t},{\bf h}_{t}|{\bf v}_{<t})= \underbrace{E_{\text{C}}({\bf v}_{t},{\bf h}_{t}|{\bf v}_{<t})}_{\text{Generative}} - \underbrace{\sum_{j,l} h_{j,t} u_{jl} y_{l,t}- \sum_{l} s_{l} y_{l,t}}_{\text{Discriminative}}$}
\end{equation}
\noindent{\bf Count-DCRBMs:} We extend the DCRBM to CountDCRBM Figure~\ref{fig:ModelFigure}. Count-DCRBMs are based on the model in \cite{Salakhutdinov_IJAR2009}, generalized to account for temporal phenomenon using CRBMs, and discriminative classification. Count-DCRBMs are used to model time varying histograms of counts. The probability distribution over the visible layer will follow a constrained Poisson distribution, $p_{\text{C-Count}}(v_{i,t}|{\bf h}_{t},{\bf v}_{<t})$ defined in (\ref{eqn:PCRBMV}), the hidden layer follows (\ref{eqn:PDCRBMH}) and the label layer follows (\ref{eqn:PDCRBMY}) and the energy function $E_{\text{C-Count}}({\bf v}_{t},{\bf h}_{t}|{\bf v}_{<t})$ defined in (\ref{eqn:EDCRBM}).
\section{Inference and Learning}\label{sec:InferenceLearning}
\noindent{\bf Inference:} to perform classification at time $t$ in the CountDCRBM given ${\bf v}_{<t}$ and ${\bf v}_{t}$ we use a bottom-up approach, computing a cost for each possible label ${\bf y}_{t}$ then choosing the label with least cost. We compute the cost for label ${\bf y}_{t}$ to be the free energy $-\log p_{\text{DC}}({\bf y}_{t},{\bf v}_{t}|{\bf v}_{<t})$ computed by marginalizing over ${\bf h}_{<t}$ and ${\bf h}_{t}$. Then, the cost associated with the candidate label is the free energy in the CountDCRBM, namely $-\log p_{\text{DC}}({\bf y}_{t},{\bf h}_{t}|{\bf h}_{<t})$ is tractable, because the sum over exponentially many terms can be algebraically eliminated.

\noindent{\bf Learning:} the parameters our model could be learned using Contrastive Divergence (CD) \cite{Hinton_NC2002}, where $\langle\cdot\rangle _{data}$ is the expectation with respect to the data distribution and $\langle\cdot\rangle _{recon}$ is the expectation with respect to the reconstructed data. The learning is done using two steps a bottom-up pass and a top-down pass using sampling equations from (\ref{eqn:PCRBMV}), (\ref{eqn:PDCRBMH}), and (\ref{eqn:PDCRBMY}). {\it Bottom-up:} the reconstruction is generated by first sampling the hidden layer $p(h_{t,j}=1|{\bf v}_{t},{\bf v}_{<t},y_l)$ for all the hidden nodes in parallel. {\it Top-down:} This is followed by sampling the visible nodes $p(v_{i,t}|{\bf h}_{t},{\bf v}_{<t})$ and $p(y_{l,t}|{\bf h}_{t}, {\bf h}_{<t})$ for all the visible nodes in parallel.
\section{Experiments}\label{sec:Experiments}
We used the open-source simulator GPGPU-Sim \cite{bakhoda2009analyzing} to generate data to validate our approach.  The simulator has been verified rigorously for accuracy against on a suite of 80 microbenchmarks \cite{leng2013gpuwattch}.  We used the BACKProp problem from the RODINIA benchmark \cite{che2010characterization}, and simulate a NVIDIA GTX480 GPU with the default configurations for GPGPU-Sim. This benchmark CUDA  program trains a feedforward neural network with one hidden layer consisting of 4096 units.

To generate our dataset, we modified GPGPU-Sim to retrieve the time-indexed list of instruction mix, ie.\! for each time cycle the number of different instruction types based on opcode. These are the visible units in our model. 
We tested our approach on three different caches (Instruction (IC), Data Read (D\_R), Data Write (D\_W))  localized within one core of the GPU. 
For each cache, GPGPU-Sim outputs a list of time-indexed binary labels corresponding to whether a cache miss occured. 
Since we want to predict a cache miss ahead of time, we aggregated the labels over $128$ cycles so a label of $y(t)=1$ means that a cache miss occurred in cycles $[t,t+128]$.

The Count-DCRBM was trained on a Tesla K20C GPU using Contrastive Divergence with a constant learning rate of $10^{-5}$. Table \ref{tab:scores_hist} shows the final accuracies of a model with $15$ hidden nodes and varying temporal history available for DCRBM. The second and third columns are metrics that describe predictive power, taking into account false positives and negatives. We observe high accuracy and predictive power of the model for all three caches. We also observe that increased history generally leads to better performance despite the increased model complexity. Our baseline is an SVM that uses the raw instruction mix as features without any temporal history. 

\begin{figure}[t]
	\centering
	\begin{minipage}{0.9\columnwidth}
		\centering
		\includegraphics[width=\textwidth]{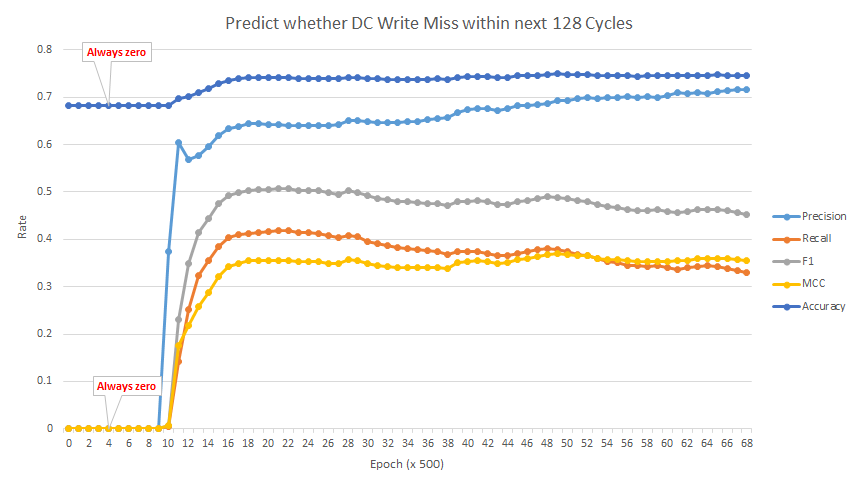}\\
		\includegraphics[width=\textwidth]{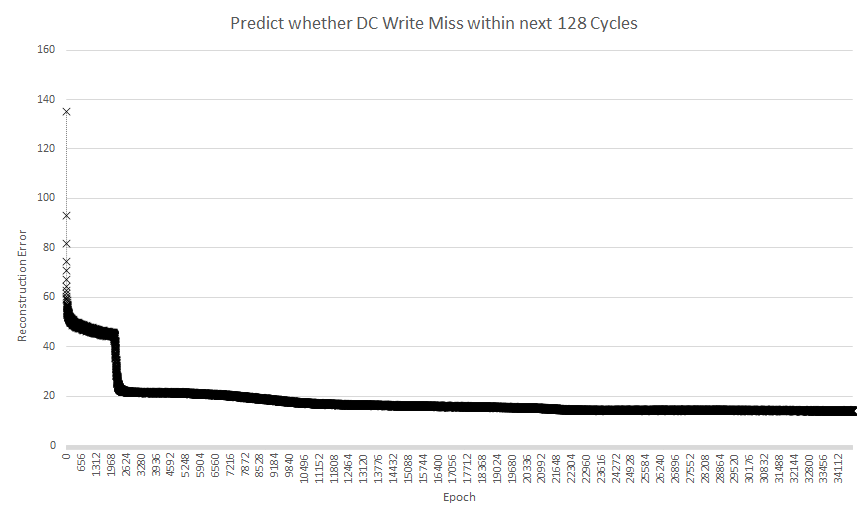}\\
		\includegraphics[width=\textwidth]{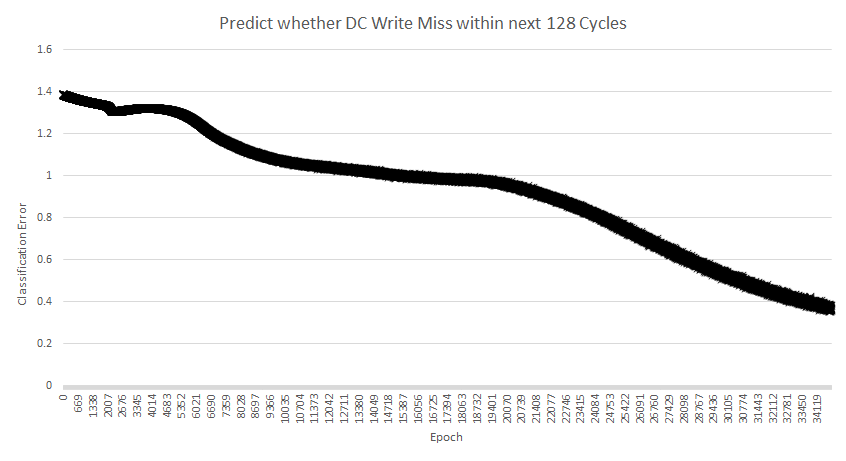}
		\caption{(Top) Accuracy (dark blue), precision light in light blue, recall in red, MCC in yellow, F1 in grey, (Middle) Reconstruction Error, (Bottom) Classification Error}
		\label{fig:Training}
	\end{minipage}
	\vfill
	\begin{minipage}{0.9\columnwidth}
		\centering
		\includegraphics[width=\textwidth]{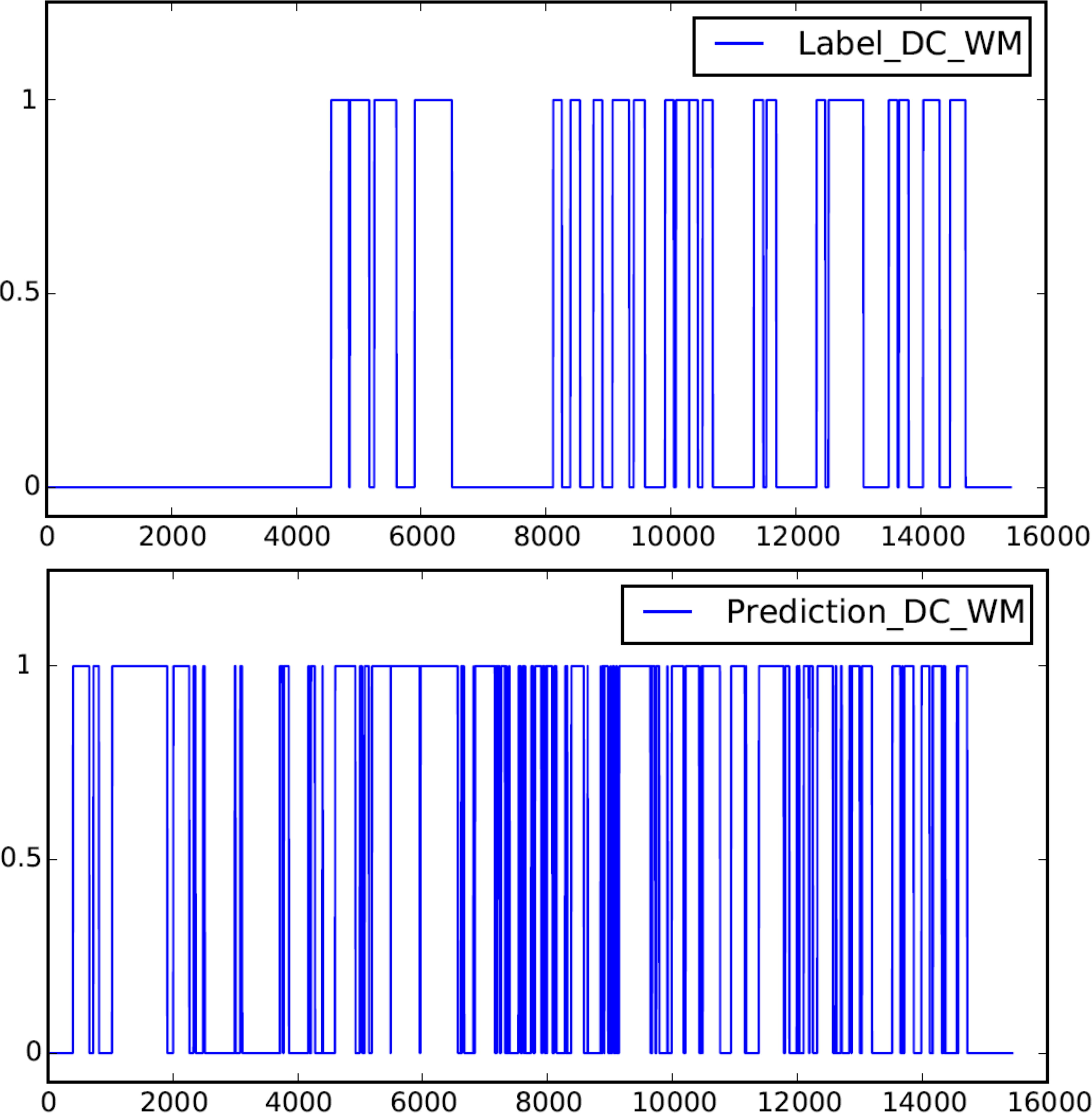}
		\caption{(Top) Ground truth labels, (Bottom) Prediction}
		\label{fig:Testing}
	\end{minipage}
\end{figure}

Model accuracy can be misleading because cache miss events are rare (e.g. about 10\% for IC). Figure \ref{fig:Training} (Top) shows these metrics over training epochs for data write cache. Note that the initial model accuracy is already about 70\% where the model predicts that cache miss never occurs, with a corresponding metric F1 and Mathew Correlation Coefficient (MCC) value of zero. As training epochs increase, we note a sharp increase in predictive power around 5000 epochs. We also show the reconstruction error in Figure \ref{fig:Training} (Middle), the objective value for training, over epochs for the data write cache.  We observe that the reconstruction error significantly drops in the first 20k epochs. Figure \ref{fig:Training} (Bottom) shows a measure of the classification error, measured in terms of the binary cross entropy between the true and predicted labels. We also observe that the classification error continues to drop steadily even though the reconstruction error has converged, showing that the model accounts for label information. Figure \ref{fig:Testing} shows the prediction using a history of 10 cycles, in comparison with the ground truth. 
Future work includes validating our approach across microbenchmarks. 
\begin{table}[t]
	\centering
	\begin{tabular}{|c|c|c|c|c|}
		\hline
		Cache&Model&MCC&F1&Accuracy\\
		\hline
		\multirow{5}{*}{DC\_R}
		&\text{DCRBM}(1)&0.32&0.59&0.67\\
		&\text{DCRBM}(5)&0.35&0.63&0.68\\
		&\text{DCRBM}(10)&{0.39}&\textbf{0.66}&0.69\\ 
		&\text{SVM(Poly)}&0.34&0.44&\textbf{0.83}\\
		&\text{SVM(RBF)}&\textbf{0.41}&0.51&\textbf{0.83}\\
		\hline
		\multirow{5}{*}{DC\_W}&\text{DCRBM}(1)&\textbf{0.36}&0.58&\textbf{0.70}\\
		&\text{DCRBM}(5)&\textbf{0.36}&\textbf{0.59}&0.65\\
		&\text{DCRBM}(10)&0.32&0.58&0.65\\
		&\text{SVM(Poly)}&0.27&0.31&0.64\\
		&\text{SVM(RBF)}&0.32&0.55&0.68\\
		\hline
		\multirow{5}{*}{IC}&\text{DCRBM}(1)&0.32&0.40&0.86\\
		&\text{DCRBM}(5)&0.32&0.39&\textbf{0.87}\\
		&\text{DCRBM}(10)&\textbf{0.37}&\textbf{0.44}&0.85\\
		&\text{SVM(Poly)}&0&0&0.99\\
		&\text{SVM(RBF)}&0&0&0.98\\
		\hline
	\end{tabular}
	\caption{Scores vs Models for different types of cache. The larger the history for DCRBM, the higher the complexity and training difficulty of the model. Larger history is better except in the case of Data Write.}
	\label{tab:scores_hist}
\end{table}
\section{Conclusions}\label{sec:Conclusion}
Our approach has significant implications for the GPU revolution of computing. A data driven approach can potentially identify mix of instructions that cause performance bottlenecks. Although we focused on cache misses, any statistic of interest to the computer architecture community such as power consumption and voltage can potentially be predicted. The extension to an online embedded setting is straightforward and could potentially save computation time. 
Prediction of performance bottlenecks is a step towards a cognitive processor architecture.
\section*{Acknowledgments}\label{sec:Acknowledgments}
This research is partially funded under NSF \#1526399, the Defense Advanced Research Projects Agency (DARPA) and the Air Force Research Laboratory (AFRL). The views, opinions and/or findings expressed are those of the authors and should not be interpreted as representing the official views or policies of the Department of Defense or the U.S. Government.
\clearpage
\bibliography{References}
\bibliographystyle{icml2016}
\end{document}